\documentclass[conference]{IEEEtran}

\usepackage{cite}
\usepackage{graphicx}
\usepackage{epstopdf}
\usepackage{amsmath}
\usepackage{algorithm}
\usepackage{algorithmic,color}
\usepackage[T1]{fontenc}

\usepackage{fancyheadings}
\begin{document}

\title{Binary Codes for Tagging X-Ray Images \\ via Deep De-Noising Autoencoders}

\author{\IEEEauthorblockN{Antonio Sze-To}
\IEEEauthorblockA{Systems Design Engineering\\
University of Waterloo\\
Waterloo, Ontario, Canada  N2L 3G1\\
Email: hy2szeto@uwaterloo.ca}
\and
\IEEEauthorblockN{H.R. Tizhoosh}
\IEEEauthorblockA{KIMIA Lab\\
University of Waterloo\\
Waterloo, Ontario, Canada  N2L 3G1\\
Email: tizhoosh@uwaterloo.ca}
\and
\IEEEauthorblockN{Andrew K.C. Wong}
\IEEEauthorblockA{Systems Design Engineering\\
University of Waterloo\\
Waterloo, Ontario, Canada  N2L 3G1\\
Email: akcwong@uwaterloo.ca}
}

\maketitle

\begin{abstract}
A Content-Based Image Retrieval (CBIR) system which identifies similar medical images based on a query image can assist clinicians for more accurate diagnosis. 
The recent CBIR research trend favors the construction and use of binary codes to represent images. 
Deep architectures could learn the non-linear relationship among image pixels adaptively, allowing the automatic learning of high-level features from raw pixels. However, most of them require class labels, which are expensive to obtain, particularly for medical images. 
The methods which do not need class labels utilize a deep autoencoder for binary hashing, but the code construction involves a specific training algorithm and an ad-hoc regularization technique. 
In this study, we explored using a deep de-noising autoencoder (DDA), with a new unsupervised training scheme using only backpropagation and dropout, to hash images into binary codes. 
We conducted experiments on more than 14,000 x-ray images. 
By using class labels only for evaluating the retrieval results, we constructed a 16-bit DDA and a 512-bit DDA independently. 
Comparing to other unsupervised methods, we succeeded to obtain the lowest total error by using the 512-bit codes for retrieval via exhaustive search, and speed up 9.27 times with the use of the 16-bit codes while keeping a comparable total error.  
We found that our new training scheme could reduce the total retrieval error significantly by 21.9\%. 
To further boost the image retrieval performance, we developed Radon Autoencoder Barcode (RABC) which are learned from the Radon projections of images using a de-noising autoencoder. 
Experimental results demonstrated its superior performance in retrieval when it was combined with DDA binary codes.
\end{abstract}

\IEEEpeerreviewmaketitle

\section{Introduction} \label{sec:introduction}
Retrieving similar medical images given a query image can be useful for more accurate diagnosis. A Content-Based Image Retrieval (CBIR) system, which identifies similar images based on an input image, is thus important for fields such as radiology and pathology.
Building such a system is formidably difficult for two reasons.
First, the similarity is hard to define for visual data.
There is a semantic gap between low-level pixel values and high-level semantics \cite{zhang2015towards}. Second, the recent advance in medical imaging devices has led to the production of a gigantic amount of image data. It has been reported that the Vanderbilt Medical Center had compiled over 100 million anonymized medical images in 18 months \cite{sklan2015toward}. A sophisticated CBIR system must be able to search through this Big Medical Image Data efficiently in response to the user's query. Hence, for medical images, accuracy and speed are the two most critical criteria for performance of a CBIR system.  

The recent trend in CBIR research is the construction and use of binary representations (codes) for image retrieval \cite{grauman2013learning}, because they offer several advantages over real-valued descriptors. For example, (1) they enable us to utilize the fast bitwise operation in hardware; (2) they are hashable, allowing the use of memory in exchange for fast retrieval; (3) they require  much less storage.

There are many methods (as reviewed in Section \ref{sec:related_work}) to hash images into binary codes. Despite their availability, they mostly seek linear projections and thus cannot capture the underlying non-linearity inherent in the image data \cite{guo2015cnn, erin2015deep}. Kernelized methods have this capability, but an appropriate kernel function, which may not exist, needs to be chosen \cite{erin2015deep}.

Recently, there are increasing attempts \cite{xia2014supervised,li2015feature,guo2015cnn,lin2015deep,zhang2015bit} to apply deep architectures (neural networks with at least 3 hidden layers), which could learn the non-linear relationship among image pixels, to hash images into binary codes. However, these methods mostly require class labels, yet obtaining class labels are expensive in the field of medical imaging. A recent study \cite{sklan2015toward} labeled 2100 medical images in collaboration with a radiologist, while the unlabeled data involved was 1 million. A very small amount of labeled data is unlikely to be adequate for training deep architectures. It may lead to over-fitting if over-trained. From the practical perspective, unsupervised methods should therefore be explored.
Semantic hashing \cite{salakhutdinov2009semantic, krizhevsky2011using} is a recent work that builds a deep autoencoder \cite{hinton2006reducing} by stacking restricted boltzmann machines (RBMs) \cite{hinton2010practical}, to learn binary codes from documents \cite{salakhutdinov2009semantic} or images \cite{krizhevsky2011using} for retrieval. 
However, building a deep autoencoder by stacking RBMs for binary hashing \cite{krizhevsky2011using} is complicated since a specific RBM training algorithm and an ad-hoc regularization technique are needed. Recently, it is reported that a deep autoencoder built by stacking de-noising autoencoders, i.e. a deep de-noising autoencoder (DDA), has comparable performance in learning features for supervised classification task \cite{vincent2010stacked}. However, the unsupervised method using a DDA to hash images into binary codes has not been explored.

Deep architectures have been applied in medical image analysis \cite{tan2011using,shin2013stacked,nayak2013classification,camlica2015autoencoding}.
There are also some methods \cite{liu2013bag, sklan2015toward} applying deep architectures for medical image retrieval but they are not hashing the medical images into binary codes, except one recent study \cite{zhang2015towards}, based on supervised methods using kernel. To our knowledge, no studies have used deep architectures to hash medical images into binary codes without class labels for retrieval.

Traditional methods seem to be unable to capture the underlying non-linearity inherent in images. Kernelized methods, on the other hand, are not adaptable. Most methods based on deep architectures require class labels. Obtaining class labels for medical images is a laborious task and not easily doable for large databases. The methods which do not need class labels train a deep autoencoder by adopting a specific training algorithm and an ad-hoc regularization technique. Until now no studies have used deep architectures to hash medical images into binary codes without class labels for retrieval. Hence, one objective of this study is to introduce a new unsupervised scheme for training a deep de-noising autoencoder (DDA) to hash images into binary codes using backpropagation and standard regularization techniques. As well, we will investigate the feasibility and performance of using a DDA to hash X-ray images into binary codes without class labels.

Indicated by our experimental results on the benchmark dataset IRMA from ImageCLEFmed09 \cite{tommasi2010overview}, the new unsupervised training scheme introduced by us had a significant impact on the retrieval performance, decreasing the total error by 21.9\%. Using the same Exhaustive Search strategy, the binary codes learned by our DDA were found to achieve the lowest total error on this dataset comparing to other unsupervised methods reported in the literature.  Using the fast retrieval strategy, we required only 7.18ms to retrieval a image on a laptop, achieving a speed-up of 9.27x over the Exhaustive Search strategy with a comparable total error.
The best unsupervised retrieval result on this dataset was achieved by the combined use of RABC and DDA binary codes.

The rest of this paper is organized as follows: The related work is reviewed in Section \ref{sec:related_work}. Our methodology is described in Section \ref{sec:methodology}. The experimental results and their analysis are shown in Section \ref{sec:results}.
The whole study is summarized and concluded in Section \ref{sec:conclusion}. 

\section{Related work} \label{sec:related_work}
\subsection{Hashing Images into Binary Codes}
The existing methods to hash images into binary codes can be categorized into two classes: data-independent and data-dependent (or learning-based) methods. Data-independent methods do not look into the data distribution when hashing images into binary codes.
One popular method in this class is Locality Sensitive Hashing (LSH) \cite{gionis1999similarity}. It uses random projections to construct hash functions such that similar images would have the same hash codes with high probability. However, it works only with theoretical guarantees for some metric spaces, e.g. $l_{0}$, $l_{2}$ and Jaccard distances \cite{wang2012semi,guo2015cnn}.

Data-depending methods investigate the data distribution when hashing images into binary codes. They are also called learning-based methods. This class of method can be categorized into two sub-classes: unsupervised and supervised. For the unsupervised sub-class, no class label is required to learn the hash functions. 
For example, Spectral Hashing \cite{weiss2009spectral} is to obtain balanced binary codes by solving a spectral graph partitioning problem, but the input data is assumed to be uniformly distributed in $R^{d}$ \cite{grauman2013learning}. Kernelized Locality-Sensitive Hashing (KLSH) \cite{kulis2009kernelized} formulates the random projections necessary for LSH in kernel space using a subset of samples, such that the underlying embedding of the data needs not to be explicitly known and computable \cite{grauman2013learning}, but an appropriate kernel is required and its scalability is questioned \cite{erin2015deep}. For the supervised sub-class, class labels are required to learn the hash functions. Labeled training data can be used for constructing a pairwise similarity matrix to learn the hashing functions. Binary Re-constructive Embedding (BRE) \cite{kulis2009learning} is to minimize the deviation between the original Euclidean distances and the learned Hamming distances.
Minimum Loss Hashing (MLH) \cite{norouzi2011minimal} is to minimize the difference between the learned Hamming distance and the binary quantization error. Kernel-based Supervised Hashing (KSH) \cite{liu2012supervised} map samples into binary codes whose Hamming distances are minimized on similar pairs and maximized on dissimilar pairs. 

Nevertheless, these approaches require the use of class labels, which are expensive to acquire if the data is large. Note that some methods only require a subset of training data having class labels.
However, they are prone to over-fitting when labeled data is small \cite{wang2012semi}. Semi-supervised hashing (SSH) \cite{wang2012semi} minimizes empirical error for the pairwise similarity in the training samples, regularized by maximizing the variance of the labeled and unlabeled data. But these methods mostly seek linear projections and thus cannot capture the non-linear structure of samples \cite{guo2015cnn, erin2015deep}. Kernelized methods could capture the underlying non-linearity, but an appropriate kernel function, which may not exist, needs to be chosen \cite{erin2015deep}. 

\subsection{Hashing by Deep Architectures}
Recently, there are increasing attempts \cite{xia2014supervised,li2015feature,guo2015cnn,lin2015deep,zhang2015bit} to use deep architecture to hash images into binary codes.
Most of these methods require class labels. They utilize Convolutional Neural Network (CNN) \cite{guo2015cnn, lin2015deep,zhang2015bit} to conduct multi-class classification, followed by binarizing the activations of a fully-connected layer with a threshold and taking the binarized results as hash codes. 
However, as mentioned before, obtaining class labels is an expensive task for large datasets, and these methods are likely to overfit if over-trained on a few labeled data \cite{wang2012semi}. On the other hand, there are limited attempts in using deep architectures to hash images into binary codes without using labels for retrieval. Semantic hashing \cite{salakhutdinov2009semantic, krizhevsky2011using} is a recent approach that builds a deep autoencoder \cite{hinton2006reducing} by stacking restricted Boltzmann machines (RBMs) \cite{hinton2010practical}, to learn binary codes from documents \cite{salakhutdinov2009semantic} or images \cite{krizhevsky2011using} for retrieval.
However, building a deep autoencoder by stacking RBMs for binary hashing \cite{krizhevsky2011using} is complicated because
1) a specific RBM training algorithm called contrastive divergence is needed, and 2) An ad-hoc regularization technique, i.e. binarizing the forward activities in the coding layer \cite{krizhevsky2011using}, needs to be adopted.
Recently, it has been reported that a deep autoencoder built by stacking de-noising autoencoders, i.e. a deep de-noising autoencoder (DDA), has a comparable performance in learning features for supervised classification task\cite{vincent2010stacked}.
To our knowledge, there are no studies on how to use a DDA to hash images into binary codes.

\subsection{Hashing Medical Images into Binary Codes}
There is a small number of studies \cite{unay2008medical,xu2009medical,nanni2010local,kanumuri2014progressive,camlica2015medical,tizhoosh2015barcode} involving hashing medical images into binary codes, and to a lesser extent for image retrieval \cite{unay2008medical,xu2009medical,tizhoosh2015barcode}. Rather than hashing the entire images into binary codes, the research community in medical image retrieval adopts localized methods such as obtaining binary descriptors from the images, assuming that some local regions are more important in medical images \cite{camlica2015medical}.
Local Binary Pattern (LBP) descriptor \cite{huang2011local} has been most commonly used for similar purposes \cite{unay2008medical,xu2009medical,tizhoosh2015barcode}.
Recently, a new descriptor called ``Radon Barcode'' was proposed  \cite{tizhoosh2015barcode,Tizhoosh2016}, which obtains binary codes from a medical image by binarizing its Radon projections. It reported superior performance than LBP in X-ray images. We compare the binary codes learned by our method with these methods, and study if it is possible to learn binary codes from Radon projections.

\section{Methodology} \label{sec:methodology}
\subsection{Problem Definition}
Let $X=\{x_{n}\}_{n=1}^{N}$ be a set of $N$ X-ray images, where each X-ray image $x_{n}$ is reshaped from a two-dimensional matrix to a one-dimensional vector. Given $X$, the problem is to learn a mapping $F:X \rightarrow \{0,1\}^{N \times k}$, i.e. to map a X-ray image $x_{n}$ to a $k$-bit binary code $b_{n} \in \{0,1\}^{k}$, such that the semantic similarity  between the images is preserved in the binary codes.

\subsection{Proposed Method}
\begin{figure*}[htb]
\begin{center}
\includegraphics[width=0.7\textwidth]{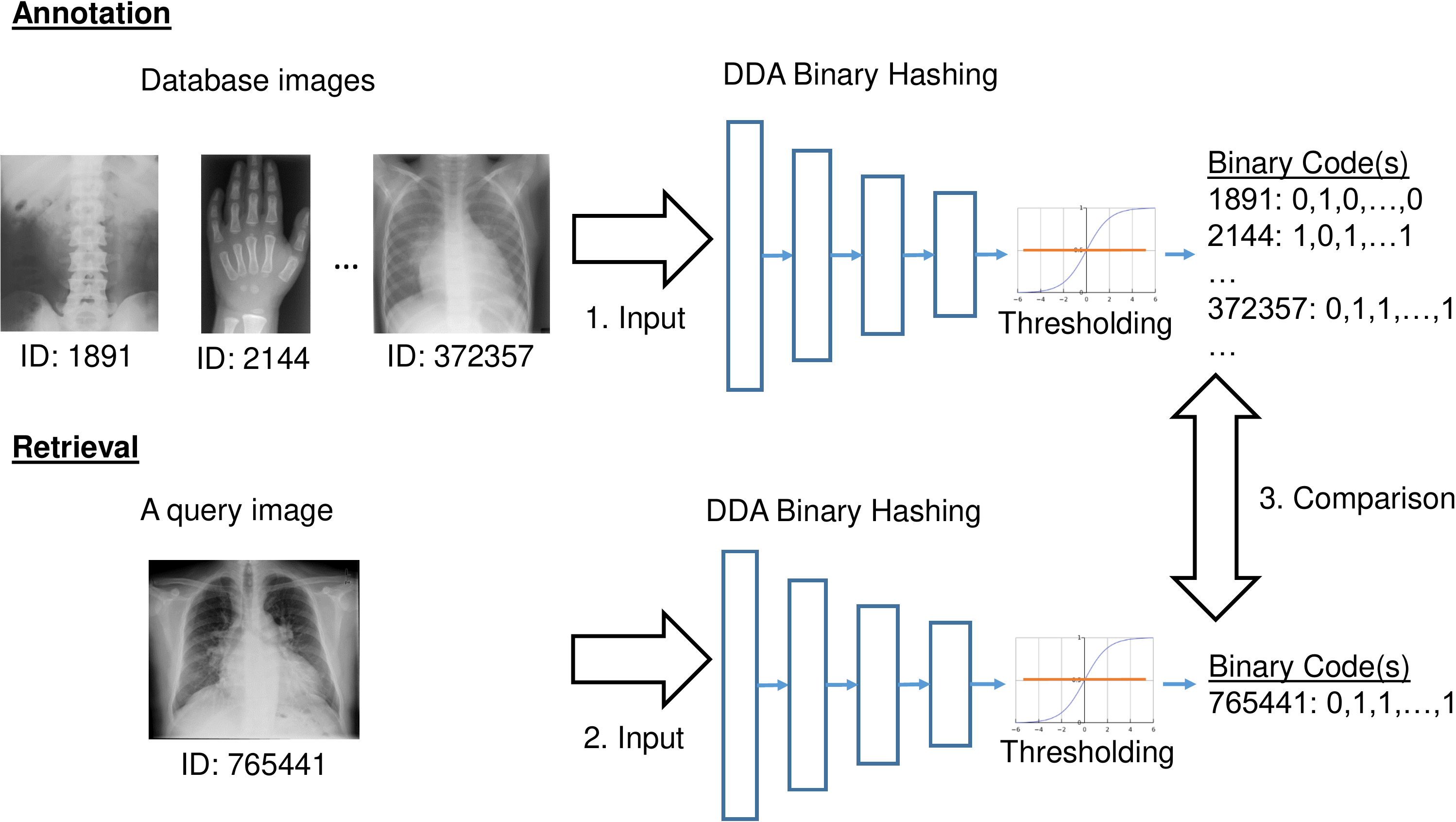}
\end{center}
\caption{Deep de-noising autoencoder (DDA). For all images, we use a trained DDA to annotate each of them with a $k$-bit binary code. Given a query image, we use the same DDA to annotate it with a $k$-bit binary code, followed by comparing this binary code with the binary codes of all database images.
}\label{fig:CBIR}
\end{figure*}

The overall idea of the proposed method is to use autoencoders to learn high-level features from X-ray images in an unsupervised manner. As well, we will use thresholding to binarize the high-level features into binary codes. How we use the binary codes for X-ray images retrieval is illustrated in Fig \ref{fig:CBIR}. Here we use a specific type of autoencoder, namely de-noising autoencoder \cite{vincent2010stacked}, to reduce learning features from noises. 
We also stack multiple de-noising autoencoders into a deep architecture called deep de-noising autoencoder (DDA) \cite{vincent2010stacked} to enhance feature learning capability. 

In general, to train a DDA for hashing X-ray images into binary codes, there are four steps:

\textbf{Step 1: Image Pre-processing.}
All images are first resized to a small size (default: 32 $\times$ 32). 
As X-ray images are in grayscale, through dividing all intensity values by 255, they are normalized to be in $[0,1]$.

\textbf{Step 2: Unsupervised Layer-by-layer Training.} (Algorithm \ref{alg:layerTrain})
A DDA is constructed by first training each layer as an individual de-noising autoencoder by backpropagation, as reported in \cite{vincent2010stacked}. 
A dropout layer \cite{srivastava2014dropout} is introduced after the input layer, such that for each training sample, a randomly chosen subset (default: 20\%) of the inputs is set to zero. Note that this drop-out layer has no effects on testing.

\begin{algorithm}[tb]
\caption{layerTrain}
\label{alg:layerTrain}
\begin{algorithmic}
\STATE \textbf{Input:} a set of training images $X$, a list of fan-in/fan-out of each layer in encoder $L_{enc}$ 
\COMMENT{e.g. $L_{enc}$ = [(1024,768),(768,512)]}  
\STATE \textbf{Output:} a list of encoder weights $W_{enc}$, a list of decoder weights $W_{dec}$
\COMMENT{ e.g. $W_{enc}$ = [$W_{enc}^{(1)}$, $W_{enc}^{(2)}$], $W_{dec}$ = [$W_{dec}^{(1)}$, $W_{dec}^{(2)}$]}
\\
\COMMENT{Constants}
\STATE epoch = 100, batchSize = 16, p = 0.2
\\
\COMMENT{Initialization}
\STATE $W_{enc}$ = [], $W_{dec}$ = []
\\
\COMMENT{Train layer by layer}
\FOR{$i$ from 1 to size($L_{enc}$)}
    \STATE encFanIn, encFanOut = $L_{enc}$[$i$]
    \STATE decFanOut, decFanIn = $L_{enc}$[$i$]
    \\
    \COMMENT{Set up a NN with a dropout layer, an input-hidden layer and a hidden-output layer, with weights initialized by the scheme in \cite{glorot2010understanding}}
    \STATE Q = setUpLayers([DropoutLayer(encFanIn, encFanIn, p), \\ SigmoidLayer(encFanIn, encFanOut), \\ SigmoidLayer(decFanIn, decFanOut)])
    \\
    \COMMENT{Train the network on $X$ for epoch times}
    \FOR{$e$ from 1 to epoch}
        \STATE trainNN(Q, $X$, batchSize) 
        \COMMENT{Train NN by backpropagation with a mini-batch of batchSize via RMSProp \cite{tieleman2012rmsprop}}
    \ENDFOR
    \\
    \COMMENT{Add the weights at the end of the weight lists}
          \STATE append($W_{enc}$, getEncoderWeights(Q));
          \STATE append($W_{dec}$, getDecoderWeights(Q));
\ENDFOR
\\
\COMMENT{Return $W_{enc}$, $W_{dec}$}
\RETURN {$W_{enc}$, $W_{dec}$}
\end{algorithmic}
\end{algorithm}

\begin{figure*}[tb]
\begin{center}
\includegraphics[width=0.7\textwidth]{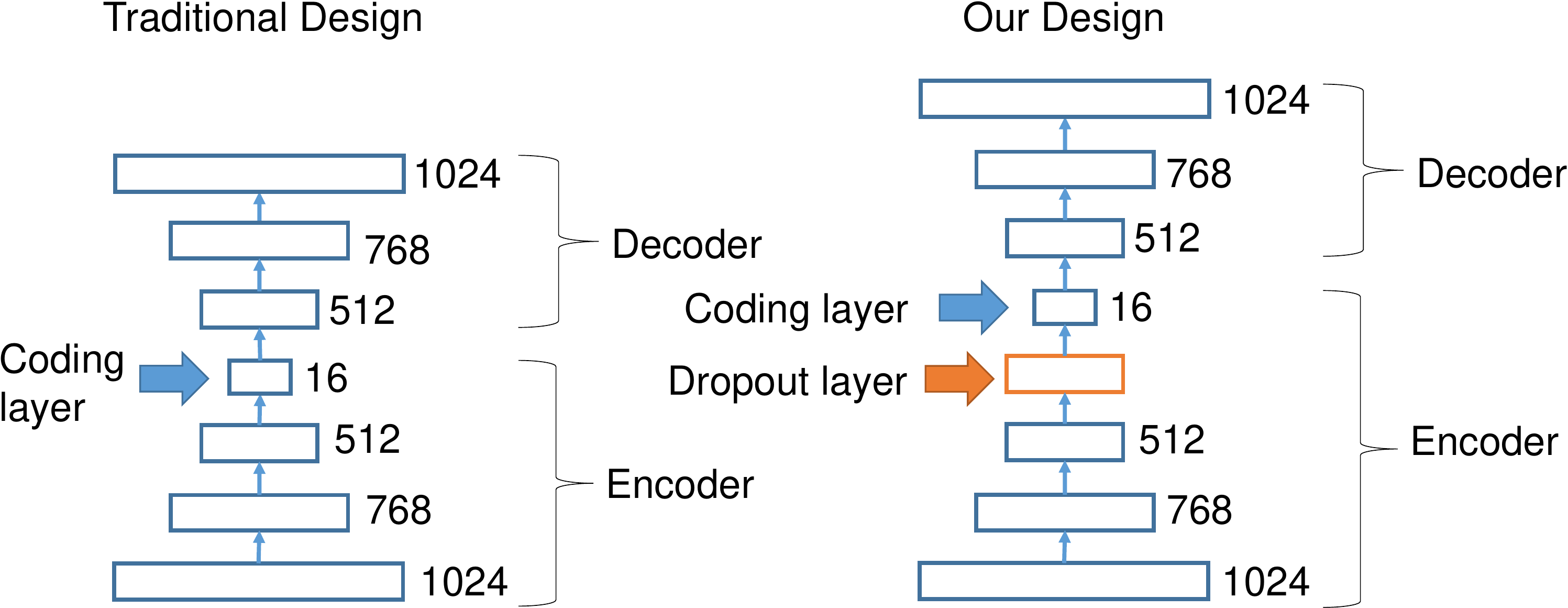}
\end{center}
\caption{This figure shows the difference between the traditional design (left) and our design (right) of a deep (de-noising) autoencoder (DDA) with the following architecture. We added a dropout layer before the coding layer to regularize it. A Dropout layer has been reported to be a way to add noise  \cite{vincent2010stacked}. Note that the dropout layer is only effective in training and has no effects in testing. We trained this entire DDA by Methodology Step 3 Unsupervised Fine-tuning.
}\label{fig:new_ae}
\end{figure*}

\textbf{Step 3: Unsupervised Fine-Tuning with Dropout.} (Algorithm \ref{alg:fineTune})
After unsupervised layer-by-layer training, the de-noising autoencoders are stacked one-by-one to construct a DDA. According to \cite{salakhutdinov2009semantic}, the last layer of the decoder is turned into a softmax layer. We introduce one new change to the architecture to improve the hashing performance. A dropout \cite{srivastava2014dropout} layer is added before the coding layer to regularize it, as shown in Fig. \ref{fig:new_ae}.
A Dropout layer has been reported to be a way to add noise \cite{vincent2010stacked}, by randomly setting the output of hidden units to be zero. Note that this dropout layer has no effects in testing. After these changes, the DDA is then trained by backpropagation. 

\begin{algorithm}
\caption{fineTune}
\label{alg:fineTune}
\begin{algorithmic}
\STATE \textbf{Input:} a set of training images $X$, a list of fan-in/fan-out of each layer in encoder $L_{enc}$, a list of encoder weights $W_{enc}$, a list of decoder weights $W_{dec}$
\COMMENT{e.g. $L_{enc}$ = [(1024,768),(768,512)], $W_{enc}$ = [$W_{enc}^{(1)}$, $W_{enc}^{(2)}$], $W_{dec}$ = [$W_{dec}^{(1)}$, $W_{dec}^{(2)}$]}  
\STATE \textbf{Output:} a trained DDA $Q$ 
\\
\COMMENT{Constants}
\STATE epoch = 100, batchSize = 16, p = 0.2
\\
\COMMENT{Initialization}
\STATE encoderLayers = [], decoderLayers = [];
\\
\COMMENT{Set up encoder layers}
\FOR{$i$ from 1 to size($L_{enc}$)}
    \STATE encFanIn, encFanOut = $L_{enc}$[i]
    \IF{i == size($L_{enc}$)}
        \STATE appendLayers(encoderLayers,
        \STATE [DropoutLayer(encFanIn, encFanIn, p)])
       \ENDIF
      \STATE appendLayers(encoderLayers, 
      \STATE [SigmoidLayer(encFanIn, encFanout, $W_{enc}$[$i$])])
\ENDFOR
\\
\COMMENT{Set up decoder layers}
\FOR{$j$ from size($L_{enc}$) downto 1}
    \STATE decFanOut, decFanIn = $L_{enc}$[j]
    \IF{j == size($L_{enc}$)}
        \STATE appendLayers(decoderLayers, 
        \STATE [SoftmaxLayer(decFanIn, decFanout, $W_{dec}$[$j$])])
    \ELSE
        \STATE appendLayers(decoderLayers, 
        \STATE [SigmoidLayer(decFanIn, decFanout, $W_{dec}$[$j$])])
    \ENDIF
\ENDFOR
\\
\COMMENT{Set up the NN}
\STATE Q = setUpLayers(merge(encoderLayers, decoderLayers) )
\\
\COMMENT{Train the network on $X$ for epoch times}
\FOR{$e$ from 1 to epoch}
    \STATE trainNN(Q, $X$, batchSize) 
    \COMMENT{Train NN by backpropagation with a mini-batch of batchSize via RMSProp \cite{tieleman2012rmsprop} for short binary codes and Adam \cite{kingma2014adam} for long binary codes}
\ENDFOR
\\
\COMMENT{Return a trained DDA}
\RETURN {$Q$}
\end{algorithmic}
\end{algorithm}

\textbf{Step 4: Decoder Removal}
After the training, the decoder in the DDA is removed. 
Then, the DDA becomes a binary hashing function for X-ray images.
To hash an image into binary codes, a normalized image, as a one-dimensional real-valued vector is fed into the trained DDA.  
After passing through all the layers,  a one-dimensional real-valued vector outputted by the last sigmoid layer of the trained DDA is obtained. 
A threshold (\textgreater 0.5) is then applied on this real-valued vector to obtain a binary code vector.

\subsection{X-ray Image Retrieval using Binary Codes}
The retrieval of X-ray images is to identify the most similar images in the databases corresponding to a user's input X-ray image. 
It should be noted that the images in the databases and the users' input images are called training images and testing images, respectively. 
The following two X-ray image retrieval strategies are studied.

\subsubsection{Exhaustive Search} 
\label{method:exhaustive}
After training a DDA, all training images are assigned with binary codes by it. For retrieval, the trained DDA is used for assigning a binary code to a given testing image. The binary code of the testing image is then compared with all binary codes of the training images. The ones with the least number of bit difference (shortest Hamming distance) are retrieved. The entire process is depicted in Fig \ref{fig:CBIR}. In our experiments, we only retrieved the first hit (the most similar image).

\subsubsection{Semantic Hashing (SH)} 
\label{method:SH}
If this strategy is adopted, two DDAs are trained, where one is for hashing X-ray images into short binary codes (16 bits), and another is for hashing X-ray images into long binary codes (512 bits). 
All training images are assigned with both short and long binary codes. 
A hash table is then used for hashing the training images using the short binary codes as hash keys. Note that multiple X-ray images can share the same hash key. 

In retrieval, the two DDAs are used to obtain both the short and long binary codes for a given test image. A number of bit flips, e.g. $H$, is then conducted on the short binary code of the test image to obtain a new short binary code. We repeat the process until all short binary codes with $H$ bit differences are obtained. Using the short binary code of the test image with the short binary codes with $H$ bit differences, a list of X-ray images are retrieved. The long binary code of the test image is then compared with those of the retrieved X-ray images. The ones with a small number of bit differences are retrieved. In our experiments, we only retrieved the first hit.

\section{Results} \label{sec:results}
\subsection{Dataset}
To evaluate the retrieval performance, we used a benchmark dataset called IRMA (Image Retrieval Medical Applications) dataset as part of ImageCLEFmed09 initiative  \cite{tommasi2010overview} which has 12677 images for training and 1733 images for testing, all images classified using IRMA codes.  
There are in total 193 classes. Each class is associated with a unique IRMA code. An IRMA code is mainly used for evaluating content-based medical image retrieval performance (hence not available in the real world). 
It is a string of 13 characters, where each of them is within the set of \{0,\ldots,9,a,\dots,z\}. The 13 characters in an IRMA code are divided into four structures, in the following format: TTTT-DDD-AAA-BBB, where T, D, A and B mean technical, directional, anatomical and biological codes respectively. 
It should be noted that these IRMA codes are classified by professionals for benchmarking. 
In the real world, no medical images have associating IRMA codes. 

\subsection{Evaluation Metric}
To evaluate the performance of image retrieval, we used the formula provided by ImageCLEFmed09 to compute the error between the IRMA codes of the testing image and the first hit retrieved training image. We then summed up the error for all testing images. 
The formula is provided as follows:
\begin{equation} \label{equation:IRMA1}
E_{Total} = \sum_{m=1}^{1733} \sum_{j=1}^{4} \sum_{i=1}^{l_{j}} \frac{1} {b_{l_{j},i}} \frac {1} {i} \delta (I_{l_{j},i}^{m}, \tilde{I}_{l_{j},i}^{m})
\end{equation}

Here, $m$ is an indicator to each image. 
$j$ is an indicator of the structure of an IRMA code.
$l_{j}$ refers to the number of characters in each structure of an IRMA code.
For example, in the IRMA code: 1121-4a0-914-700, $l_{1}=4$, $l_{2}=3$, $l_{3}=3$ and $l_{4}=3$. 
$i$ is an indicator to a character in a particular structure. 
Here, $l_{2,2}$ refers to the character ‘a’ and $l_{4,1}$ refers to the character ‘7’. 
$b_{l_{j},i}$ refers to the number of branches, i.e. number of possible characters, at the position $i$ in the $l_{j}^{th}$ structure in an IRMA code. 
$I^{m}$ refers to the $m^{th}$ testing image and $\tilde{I}^{m}$ refers to its top 1 retrieved image. 
$\delta (I_{l_{j},i}^{m}, \tilde{I}_{l_{j},i}^{m})$ compares a particular position in the IRMA code of the testing image and the retrieved image. 
It then outputs a value in \{0, 1\} according to the following rules.

\begin{equation} \label{equation:IRMA2}
\delta (I_{l_{j},i}^{m}, \tilde{I}_{l_{j},i}^{m})=
\begin{cases}
0, & I_{l_{j},h}^{m} = \tilde{I}_{l_{j},h}^{m} \forall h \leq i \\
1, & I_{l_{j},h}^{m} \neq \tilde{I}_{l_{j},h}^{m} \exists h \leq i
\end{cases}
\end{equation}

We used the Python implementation of the above formula provided by ImageCLEFmed09 to compute the errors. 

\subsection{Implementation and Parameter Setting}
The deep learning library Keras (http://keras.io/) with Theano backend \cite{bastien2012theano} is adopted for implementation. The parameter setting is described as follows: 
The dropout parameter was set as 0.2, i.e. 20\% of the inputs would be randomly set as zeroes. For both Methodology Step 2 and Step 3, the number of epochs and batch size were 100 and 16, respectively. 
The default loss function was binary cross-entropy. The default optimizer in Methodology Step 2 was RMSProp  \cite{tieleman2012rmsprop}.
The default optimizer in Methodology Step 3 was RMSProp \cite{tieleman2012rmsprop} for short codes and Adam \cite{kingma2014adam} for long codes.
These optimizers were used with default settings.
All the experiments were run on a computer with 8.0 GB RAM, a i5-2410M-2.30GHz CPU (4 Cores) and a GT520M Graphics card.
The neural networks were trained on the GPU. These settings were used in all experiments unless further specified.

\subsection{Experiment Series 1}
The objective of this experiment to investigate whether the binary codes tagged by a DDA on X-ray images can be used for image retrieval without using class labels. First, we pre-processed the training and testing images, according to Step 1 in the Methodology section. Second, according to Steps 2 and 3, we trained a DDA on the training images. Third, using Step 4, we used them to tag all training and testing images with binary codes. 
Afterward, we studied the X-ray image retrieval performance using these binary codes, according to an exhaustive search strategy mentioned in Section \ref{method:exhaustive}. 
We evaluated the performance using Equations \ref{equation:IRMA1} and \ref{equation:IRMA2}. 
The results are shown in Table \ref{tab:Experiment1}.
There are two types of DDA, where one is for short binary codes (16 bits) with the encoder architecture: 1024 inputs $\rightarrow$ 768 sigmoid neurons $\rightarrow$ 512 sigmoid neurons $\rightarrow$ Dropout $\rightarrow$ 16 sigmoid neurons and another is for long binary codes (512 bits) with the architecture: 1024 inputs $\rightarrow$ 768 sigmoid neurons $\rightarrow$ Dropout $\rightarrow$ 512 sigmoid neurons. 
Note that the dropout layer is only useful in training but not in testing. For comparison, the X-ray image retrieval errors of other binary codes such as Radon Barcode (RBC) \cite{tizhoosh2015barcode}, Local Binary Pattern (LBP) \cite{huang2011local} and Local Radon Binary Pattern (LRBP) \cite{galoogahi2012face} on the same IRMA dataset using the Exhaustive Search strategy are listed from \cite{tizhoosh2015barcode} for reference.

\begin{table}[htb]
\centering
\caption{A comparison of the image retrieval performance between the binary codes learned from Deep De-noising Autoencoder (DDA) and other methods for IRMA images.}
\begin{tabular}{cccc} \hline
    Binary Code / Method & Label needed & Length of code & $E_{Total}$   \\ \hline
    TAUbiomed \cite{avni2009addressing} & Yes & N/A & 169.5 \\ 
     $DDA_{1024 \rightarrow 768 \rightarrow 512 \rightarrow 16}$ & No & 16 & 703.95 \\ 
    $DDA_{1024 \rightarrow 768 \rightarrow 512}$ & \textbf{No} & 512 & \textbf{344.08}  \\ 
    $RBC_{4}$ \cite{tizhoosh2015barcode} & No & 512 & 476.62  \\ 
    $RBC_{8}$ \cite{tizhoosh2015barcode} & No & 1024 & 478.54  \\ 
    $RBC_{16}$ \cite{tizhoosh2015barcode} & No & 2048 & 470.57  \\ 
    $RBC_{32}$ \cite{tizhoosh2015barcode} & No & 4096 & 475.62  \\ 
    $LBP$ \cite{huang2011local} & No & 7200 & 463.81  \\ 
    $LRBP_{4}$ \cite{galoogahi2012face} & No & 7200 & 483.54  \\ 
    $LRBP_{32}$ \cite{galoogahi2012face} & No & 7200 & 501.96  \\ 
     \hline
\end{tabular}
\label{tab:Experiment1}
\end{table}

As shown in Table \ref{tab:Experiment1}, the 512-bit binary code learned by DDA has achieved a lower error total ($E_{Total}$) comparing to all the other binary codes including the latest developed Radon Barcode \cite{tizhoosh2015barcode}.  
We also observe that even the length of binary codes has increased, the error total is still much higher compared to the 512-bit binary code. 
Note that the lowest error total achieved in this dataset is 169.5 by TAUbiomed \cite{avni2009addressing} which requires class labels.
No class information was used in ours and the rest in Table \ref{tab:Experiment1}. 

\subsection{Experiment Series 2}
The objective of this experiment was to investigate whether Semantic Hashing (SH) (Section \ref{method:SH}) can be applied on the X-ray images to speed up the retrieval process. 
We used the 16-bit and 512-bit DDA binary codes for hashing and re-ranking, respectively. 
The number of bit flips was to be 1, 2 and 3. 
For comparison, we also recorded the error total and retrieval time per image for a baseline Pearson Correlation method.
Given a test image, the Pearson Correlation Coefficient was computed between the test image and every training image. 
The training image associated with the highest absolute Pearson Correlation Coefficient was retrieved.
This baseline Pearson Correlation method was studied on both 32 $\times$ 32 and 64 $\times$ 64 images. 
The results are shown in Table \ref{tab:Experiment2}.

\begin{table*}[htb]
\centering
\caption{A comparison of the image retrieval performance between Exhaustive Search Strategy, Semantic Hashing (SH) and Pearson Correlation for IRMA images.}
\begin{tabular}{ccccc} \hline
    Binary code / method & Length of code & Training time (s) & Retrieval time (ms) per image & $E_{Total}$   \\ \hline
     $DDA_{1024 \rightarrow 768 \rightarrow 512 \rightarrow 16}$ & 16 & 5420.09$\pm$10.65 & 50.93$\pm$0.86 & 703.95 \\ 
    $DDA_{1024 \rightarrow 768 \rightarrow 512}$ & 512 & 5084.72$\pm$6.13 &66.58$\pm$1.22 & 344.08 \\ 
    SH (Number of Bit Flips: 1) & 16; 512 & 10504.81$\pm$14.02 & 5.62$\pm$0.11 & 414.80 \\ 
    SH (Number of Bit Flips: 2) & 16; 512 & 10504.81$\pm$14.02 & \textbf{7.18$\pm$0.12} & \textbf{378.30} \\ 
    SH (Number of Bit Flips: 3) & 16; 512 & 10504.81$\pm$14.02 & 24.91$\pm$0.76 & 365.24 \\ 
    Pearson Correlation (32 $\times$ 32) & N/A & N/A & 2311.60$\pm$37.89 & 399.68 \\ 
    Pearson Correlation (64 $\times$ 64) & N/A & N/A & 2587.91$\pm$30.34 & 402.05  \\ 
     \hline
\end{tabular}
\label{tab:Experiment2}
\end{table*}

The Table \ref{tab:Experiment2} records the training time, error total and retrieval time per image for different methods.
Note that the training time and the retrieval time per image were the averages of 20 independent runs of the full testing set (starting with the same random seed), along with the 95\% confidence interval. 
As shown, Semantic Hashing with 2-bits flip has achieved a speedup of 9.27x over the Exhaustive Search Strategy via 512-bit binary codes learned from a DDA with a comparable total error, and a speed-up of 411.32x over the baseline Pearson Correlation method with a lower error total. 

\subsection{Experiment Series 3}
The objective of this experiment was to investigate whether we can (1) learn binary codes from Radon projections \cite{tizhoosh2015barcode} of X-ray images and (2) use them for boosting the image retrieval performance. 
We first computed the Radon projections of all training and test images with the size of 256 $\times$ 256 on 16 projections such that each image has a 4096-dimension real-valued vector. To obtain the binary code, it was proposed in \cite{tizhoosh2015barcode} to use the median value of each projection to binarize the Radon projections, known as Radon Barcode (RBC) \cite{tizhoosh2015barcode}.
In this experiment, we constructed a de-noising autoencoder with the encoder architecture: 4096 inputs $\rightarrow$ Dropout $\rightarrow$ 2048 sigmoid neurons.
We trained this de-noising autoencoder on the Radon projections using the default parameter setting, except that in Methodology Step 3 we set the epoch to be 2200. We name the binary code learned as Radon Autoencoder Barcode (RABC). We further define the bit difference between a test image $I^{m}$ and a candidate image $I^{c}$ as $\frac{d_{RABC2048}(I^{m},I^{c})}{2048} + \frac{d_{DDA512}(I^{m},I^{c})}{512}$,
where the former is the normalized bit difference between the 2048-bit RABC of the testing and the candidate image,
and the latter is the normalized bit difference between the 512-bit DDA binary code of the testing and the candidate image.
The image retrieval results under the Exhaustive Search strategy are shown in Table \ref{tab:Experiment3}.

\begin{table*}[htb]
\centering
\caption{Comparison of image retrieval performance for IRMA images.}
\begin{tabular}{ccccc} \hline
    Binary code / method & Length of code & Training time (s) & Retrieval time (ms) per image & $E_{Total}$   \\ \hline
    $DDA_{1024 \rightarrow 768 \rightarrow 512}$ & 512 & 5084.72$\pm$6.13 &66.58$\pm$1.22 & 344.08 \\ 
    $RBC_{256\times256, 16 proj}$ & 4096 & N/A & 162.50$\pm$0.37 & 546.09 \\ 
    $RABC_{4096 \rightarrow 2048}$ & 2048 & about 270,600 & 109.28$\pm$0.29 & 362.54 \\ 
    $DDA_{1024 \rightarrow 768 \rightarrow 512}$/512 + $RABC_{4096 \rightarrow 2048}/2048$ & 512;2048 & about 275,690 & 233.30$\pm$2.88 & \textbf{330.60} \\ 
     \hline
\end{tabular}
\label{tab:Experiment3}
\end{table*}

\begin{figure}[htb]
\begin{center}
\includegraphics[width=0.4\textwidth]{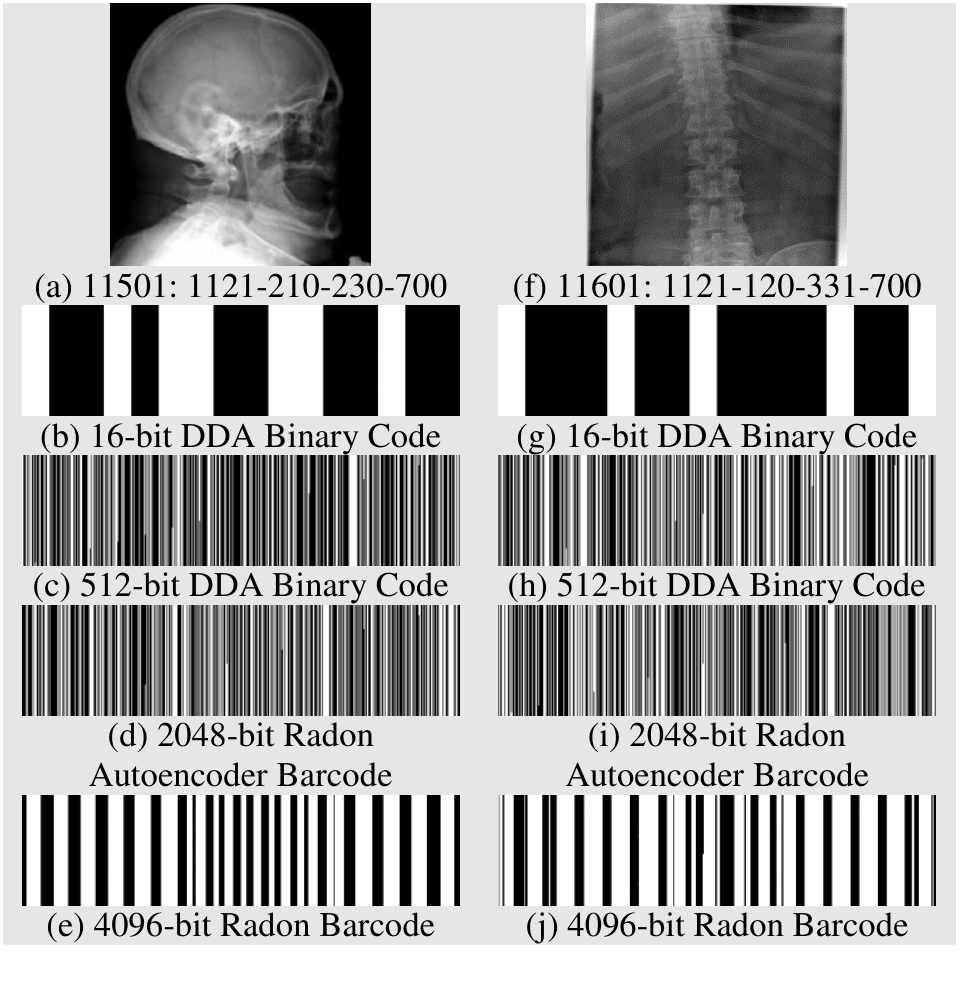}
\end{center}
\caption{This figure shows the IRMA code, 16-bit DDA Binary Code, 512-bit DDA Binary Code, 2048-bit Radon Autoencoder Barcode and 4096-bit Radon Barcode of image 11501 (a-e) and 11601 (f-j).
}\label{fig:barcode}
\end{figure}

As shown, the error total achieved by RABC is significantly lower than that of RBC, even though the length of binary code of former is half of the latter.
This shows that the binary codes learned from the Radon projections contain much compact information.
The best image retrieval performance was achieved by the combined use of RABC and the 512-bit DDA binary codes learned in Experiment 1. 
This shows the effectiveness of RABC in improving image retrieval performance. A demonstration of binary codes is in Fig. \ref{fig:barcode}.

\subsection{Experiment Series 4}
The objective of this experiment was to investigate the effects brought by the dropout layer on the binary hashing performance of a DDA. In Methodology Step 3 (Unsupervised Fine-tuning), we introduced a change in the DDA architecture that is different from both the DDA mentioned in \cite{vincent2010stacked} and the very deep autoencoder mentioned in \cite{krizhevsky2011using}. 
We added a dropout \cite{srivastava2014dropout} layer before the coding layer, where 20\% randomly chosen inputs to the coding layer would be set as zeros in training. We also used different optimizers (Adam \cite{kingma2014adam} or RMSProp \cite{tieleman2012rmsprop}) in Methodology Step 3 to identify the best training scheme.

We focused our experiment on the following DDA encoder architecture: 1024 inputs $\rightarrow$ 768 sigmoid neurons $\rightarrow$ Dropout $\rightarrow$ 512 sigmoid neurons.
Based on this configuration, for each possible scenario, we started from the same set weights initialized by Methodology Step 2 and trained the DDA using Methodology Step 3.
After learning the binary codes, we studied the image retrieval performance under the Exhaustive Search Strategy. 
The results are in Table \ref{tab:Experiment4}.

\begin{table}[htb]
\centering
\caption{Comparison of image retrieval performance}
\begin{tabular}{ccc} \hline
    Configuration  & Optimizer & $E_{Total}$   \\ \hline
    No Step 3 & N/A & 440.40 \\ \hline
    Step 3 without dropout & RMSProp & 410.64 \\
     Step 3 & RMSProp & \textbf{370.70} \\ \hline
    Step 3 without dropout & Adam & 561.47 \\
     Step 3 & Adam & \textbf{344.08} \\     
     
     \hline
\end{tabular}
\label{tab:Experiment4}
\end{table}

We observe that the dropout layer brought to the DDA has a significant improvement effect on the binary hashing performance.
In Table \ref{tab:Experiment3}, the error total with no step 3 is 440.40. Using RMSProp as the optimizer, if we use Step 3 without dropout, the error is decreased by 29.76. If we use Step 3 (with dropout), the error is further decreased by 39.94.
Using Adam as the optimizer, if we use Step 3 without dropout, the error is increased by 121.07, probably because of over-fitting.
If we use Step 3 (with dropout), the error is decreased by 96.32, a reduction of total error by 21.9\%. This indicates the importance of adding a dropout layer before the coding layer, independent of the optimizer used.

\section{Conclusion} \label{sec:conclusion}
In this study, we explored the use of Deep De-noising Autoencoder (DDA) to hash X-ray images into binary codes without class labels. 
We introduced a new unsupervised training scheme by adding a dropout layer to the DDA architecture in the step of Unsupervised Fine-tuning. Conducting experiments on the benchmark dataset IRMA from ImageCLEFmed09, we observe that this is important for satisfactory binary hashing performance, reducing the total retrieval error by up to 21.9\%. Moreover, we demonstrated an alternative in order to construct a deep (de-noising) autoencoder by stacking (de-noising) autoencoder directly for binary hashing, using only backpropagation and dropout, simplifying the implementation. Furthermore, we developed Radon Autoencoder Barcode (RABC) and used it to improve retrieval performance. All these indicate the potential of our method for CBIR in practice, specifically for big image data like medical images whose labels are expensive to obtain.

\bibliographystyle{IEEEtran}

\bibliography{IEEEfull}

\end{document}